\documentclass[runningheads]{llncs}
\usepackage[T1]{fontenc}
\usepackage{graphicx}
\usepackage{multirow}
\usepackage{rotating}
\usepackage{subcaption}

\begin{document}
%
\title{Comprehensive Equity Index (CEI): Definition and Application to Bias Evaluation in Biometrics}

\titlerunning{Comprehensive Equity Index (CEI)}
%



\author{Imanol Solano\inst{1,2} \and
Alejandro Peña\inst{1} \and Aythami Morales\inst{1} \and Julian Fierrez\inst{1} \and
Ruben Tolosana\inst{1} \and
Francisco Zamora-Martinez\inst{2} \and
Javier San Agustin\inst{2}}

\authorrunning{I. Solano et al.}

%
\institute{BiDA-Lab, Universidad Autónoma de Madrid, 28049, Madrid, Spain\\
\email{\{alejandro.penna, aythami.morales, julian.fierrez, ruben.tolosana\}@uam.es}\and
Veridas, 31192, Navarre, Spain\\
\email{\{isolano, pzamora, jsanagustin\}@veridas.com}}
%
\maketitle              
\begin{abstract}
We present a novel metric designed, among other applications, to quantify biased behaviors of machine learning models. As its core, the metric consists of a new similarity metric between score distributions that balances both their general shapes and tails' probabilities. In that sense, our proposed metric may be useful in many application areas. Here we focus on and apply it to the operational evaluation of face recognition systems, with special attention to quantifying demographic biases; an application where our metric is especially useful. The topic of demographic bias and fairness in biometric recognition systems has gained major attention in recent years. The usage of these systems has spread in society, raising concerns about the extent to which these systems treat different population groups. A relevant step to prevent and mitigate demographic biases is first to detect and quantify them. Traditionally, two approaches have been studied to quantify differences between population groups in machine learning literature: 1) measuring differences in error rates, and 2) measuring differences in recognition score distributions. Our proposed Comprehensive Equity Index (CEI) trade-offs both approaches combining both errors from distribution tails and general distribution shapes. This new metric is well suited to real-world scenarios, as measured on NIST FRVT evaluations, involving high-performance systems and realistic face databases including a wide range of covariates and demographic groups. We first show the limitations of existing metrics to correctly assess the presence of biases in realistic setups and then propose our new metric to tackle these limitations. We tested the proposed metric with two state-of-the-art models and four widely used databases, showing its capacity to overcome the main flaws of previous bias metrics.


\end{abstract}
\section{Introduction}
\label{sec:intro}

In the past decade, we have experienced a revolution in the field of Artificial Intelligence (AI). The surprising capabilities of data-driven automatic systems have made possible the development of AI-based solutions in a variety of domains, such as health~\cite{gomez2021improving}, education~\cite{daza2023edbb}, or recruitment~\cite{2023_SNCS_Human-Centric_Pena}. Among these application areas, biometric recognition technology, or biometrics, is growing significantly due to its advantages over traditional security/authentication approaches. Compared to the latter, biometric recognition systems are capable of authenticating the identity of a person using features extracted from biometric data of the individual, usually known as biometric traits~\cite{jain2011introduction}. Some traits that have been studied in the field include iris~\cite{alonsofernandez2024periocular}, fingerprint~\cite{2023_IJCB_LivDet2023_Marco}, or human-computer interaction signals~\cite{Delgado2024swipe}. Of the different traits traditionally addressed in biometrics, Face Recognition (FR) is probably the one that has benefited the most from the emergence of Deep Learning. The success of novel architectures~\cite{He2015DeepRL} and learning strategies~\cite{Deng2018ArcFaceAA,Wang2018CosFaceLM}, has notably raised the performance of these systems compared to traditional handcrafted approaches~\cite{2018_IntelligentSystems_icb-rw}. Consequently, during the last decade a lot of attention has been paid to face recognition~\cite{wang2021deep}, while its use in real systems has become more widespread, with applications ranging from border control~\cite{2023_IET-Biom_Schengen_Busch} to mobile phone authentication~\cite{2019_HBookSelfie_SuperSelfieFaceIris_Alonso,2020_SSIMbook_QID_Perera}.

Although several AI-based systems may appear to be ready for large-scale deployment in a vast array of domains and applications attending solely to the performance, some unsolved issues around their use must be first addressed for successful and trustworthy applications~\cite{herrera23trust,2023_SNCS_Human-Centric_Pena}. Several scholars have raised concerns about aspects such as vulnerabilities to attacks~\cite{ghafourian2023saflsybilawarefederatedlearning,2023_Book-PAD_Face_JHO,2021_ICIP_EmoVulnerable_Pena,ramachandra2017presentation}, or potential algorithmic discrimination effects~\cite{mehrabi2021survey,2022_AI_SensitiveLoss_IS}. Attending to the latter, in recent years we have observed a significant amount of systems exhibiting biased behaviors, leading to unfair treatment towards certain individuals based on their membership to demographic groups~\cite{Buolamwini2018Gender}. In addition to these concerns, biometrics has been singled out in both past and future legislation on data privacy and AI, due to the sensitive nature of the data it deals with. The European General Data Protection Regulation (GDPR)\footnote{\url{https://gdpr-info.eu/}} already imposed several restrictions on how to store and process personal data~\cite{goodman2016step}. Furthermore, the recently approved European AI Act\footnote{\url{https://www.euaiact.com/}} includes new requirements that biometric systems shall meet, including the prevention of the aforementioned problems. Biometrics has been particularly fruitful in this scenario, with the study of demographic biases, including their prevention and mitigation~\cite{drozdowski2020demographic,deFreitasPereira2020FairnessIB}. Methods tackling bias in biometrics span the evaluation of trained models in different populations~\cite{alcala2023measuring,2022_SafeAI_IFBiD_Serna,serna2021insidebias}, learning strategies including fairness constraints in their optimization objectives~\cite{Gong2019JointlyDF,2022_AI_SensitiveLoss_IS}, or the development of new databases with a broad and fair representation of the demographic groups~\cite{2021_TPAMI_SensitiveNets_Morales,wang2020mitigating,Wang2018RacialFI}. However, a key point for analyzing and mitigating demographic biases is to be able to measure them properly.

Traditionally, the measurement of demographic bias in biometric systems has been based on the differences among the error rates for each demographic group. This approach has received the name of the differential outcome~\cite{howard2019effect}. In ~\cite{deFreitasPereira2020FairnessIB} the authors propose to measure fairness using the maximum difference (i.e., worst-case differentials) of both False Match Rate (FMR) and False Non-Match Rate (FNMR) between demographic groups at a given operating threshold $\tau$. They combine both measures in a single metric, known as Fairness Discrepancy Rate (FDR). In ~\cite{Howard2022EvaluatingPF} the authors introduced the Gini Aggregation Rate for Biometric Equitability (GARBE) metric, a fairness measure inspired by the Gini coefficient, computed for both FMR and FNMR. The National Institute of Standards and Technology (NIST) has also highlighted the use of FMR/FNMR discrepancies as a quantitative measure of system fairness~\cite{grother2019face}. They proposed the Inequity metric (IN), an alternative measure to the FDR~\cite{grother2022face}. In this case, instead of computing the maximum difference of FMR/FNMR across groups, the maximum ratios of these between demographic groups at a given threshold $\tau$ are obtained, which are combined to obtain the Inequity metric. The metric has a direct operational sense since it directly represents the number of times that the individuals from the disfavoured group are more likely to be confused, compared to the favored group (i.e., the one with the best performance). A common drawback of all the previous proposals is the need to select a concrete operational point to measure fairness, which may hinder a complete assessment of the model performance. An exception to the aforementioned threshold-focused approach that still is based on differential outcomes is the work proposed by Gong \textit{et al.}~\cite{Gong2019JointlyDF}. They proposed to measure bias as the standard deviation of performance across demographic groups, reported in terms of the Area Under the Curve (AUC). Whilst it does not require explicitly fixing a threshold, the metric cannot measure fairness at the distribution level.


Contrary to the aforementioned differential outcome approaches, Kotwal and Marcel focused on a differential performance approach~\cite{kotwal2023fairness}. Instead of measuring the fairness at a specific operating point $\tau$, they introduced a metric directly working with score distributions of demographic groups. Concretely, the Distribution Fairness Index (DFI) measures the difference in score distributions between demographic groups by leveraging the Kullback-Leibler (KL) divergence. While measuring fairness at the score distribution level poses some benefits compared to other methods, the DFI exhibits some limitations. By considering the whole distribution equally, biases appearing in the tail of those distributions, which ultimately condition the recognition performance, are not always properly represented. This nuance is especially relevant for high-performance systems, such as the ones leading the NIST's Face Recognition Technology Evaluation (FRTE). These FR systems, mostly commercial from the industry, usually present low error rates, and hence their differences are determined by extremely competitive operational points. In other words, it is mostly in the matching score distributions tails where the main differences across systems arise. The differential outcome approach can be useful to measure bias in these scenarios. Still, it should be noted that performance differences in the distribution tails may not be always related to demographic biases. As Therhörst \textit{et al} showed, non-demographic attributes such as head-pose, illuminations, brightness, resolution, or even black and white images can affect the performance of FR systems~\cite{terhorst2022comprehensive}. Therefore, it would be desirable to use a metric that strikes a balance between the two extremes: based on decision thresholds or full score distributions.

In this work, we propose a metric that can capture differences in the distributions tails without fixing specific operational points. We were inspired by recent fairness metrics~\cite{grother2022face,kotwal2023fairness} and we have addressed some of the flaws observed when using these metrics in real and synthetic scenarios~\cite{melzi2024frcsyn,shahreza2024sdfr}. We present a modification of the DFI metric~\cite{kotwal2023fairness} that solves the previously mentioned drawbacks, which we use to evaluate a high-performance algorithm presented to the NIST FRTE challenge in several state-of-the-art face recognition databases. The main contributions of this work are:

\begin{enumerate}
    \item We present an evaluation of six fairness metrics applied to face recognition in synthetic and real-world scenarios. We characterize its performance when measuring the fairness of commercial algorithms in state-of-the-art datasets. Our experiments demonstrate the difficulties in detecting biased behaviors in high-performance algorithms characterized by long-tail score distributions.  
    \item We propose a new metric, the Comprehensive Equity Index (CEI) that addresses the drawbacks of the existing DFI metric when evaluating high-performance systems. The proposed metric aims to detect demographic biases in the distributions tails and in the center part of the distributions.
\end{enumerate}


\section{Measuring Fairness in Biometric Systems}
\label{sec:measuring}

\subsection{Problem Formulation}
\label{sec:problem}

Let us consider any 2-class classification problem (n-class can be developed as multiple 2-class problems), in our case exemplified using Face Recognition (FR). Other AI setups apart from classification in which output probabilities for different data populations can be obtained are also easily covered by our methods. Other AI setups in which class probabilities are not straightforward, e.g., regression, will need further work for our ideas to be properly developed. Let's now focus for concreteness and without loss of generality in Face Recognition. 

Traditionally, a FR system operates in one of the following setups: \textit{i)} Identification or \textit{ii)} Verification. Our interest here is in the latter, where the task is to determine whether two samples belong to the same identity or not with a $1$:$1$ similarity comparison, or \textit{match}. If both samples belong to the same identity, the samples are said to form a \textit{genuine} pair, otherwise, we refer to it as \textit{impostor}, i.e., a 2-class classification problem. In a real scenario, it is common that one of the biometric samples in the pair is pre-enrolled in the system (reference sample), whose identity is known. Thus, the system is presented with a second sample (probe sample) that claims to belong to the same identity as the reference. 

Formally, to measure the performance of the system let us consider a dataset of biometric samples, which contains $N$ samples, i.e., face images $\mathbf{I}$ in the case of FR systems. Each of the images was captured from a subject, who is part of a demographic group $d_i$ according to its demographic traits (e.g., gender, ethnicity, age). We assume here a set $\mathcal{D}$ of $K$ demographic groups, which are disjoint (i.e., a subject can only be a member of one group). A FR model $\mathbf{w}^F$ is trained to extract face representations $\mathbf{x} = f(\mathbf{I}|\mathbf{w}^F)$ discriminant for the task of identity recognition from the images. The dissimilarity of a pair of face representations $(\mathbf{x}_n, \mathbf{x}_m)$ is usually computed using a vector distance metric, such as the Euclidean or the cosine distances, which can then be normalized~\cite{Fierrez-Aguilar2005_ScoreNormalization} to a score $s(\mathbf{x}_n, \mathbf{x}_m)$. In this context, different considerations of what is meant for the system to be demographically fair have been proposed in the literature~\cite{deFreitasPereira2020FairnessIB,hardt2016equality,kotwal2023fairness,2022_AI_SensitiveLoss_IS}, but they all follow a similar hypothesis: the ``performance'' of the system should be equal across groups. The nuance here is how to define performance. If we abstractly formulate the performance of a model for a demographic group $d_i$ as $\textrm{Performance}(\mathbf{w}^F|d_i)$, the previous idea would be satisfied if $\textrm{Performance}(\mathbf{w}^F|d_i) \approx \textrm{Performance}(\mathbf{w}^F|d_j)\,\, \forall d_i, d_j \in \mathcal{D}$ (where $\approx$ should be carefully defined). In traditional Machine Learning~\cite{hardt2016equality}, these performances have been measured as the probability of a certain decision of the model (i.e., Demographic Parity), or even as the True Positive Rate (i.e., Equal of Opportunity). Particularly in biometrics, differential outcome approaches have considered error rates, i.e., FMR/FNMR, as the basis for these performance measurements~\cite{deFreitasPereira2020FairnessIB}. On the other hand, differential performance approaches such as~\cite{kotwal2023fairness} consider the entire distributions of scores $z=p(s|\mathbf{w}^F)$ (where $p$ denotes probability) to represent the performance of the model.

\subsection{Fairness Metrics: Existing Methods}
\label{sec:metrics}

Recently, Kotwal and Marcel have addressed the problem of measuring demographic fairness in biometric systems~\cite{kotwal2023fairness}. They argued how the community has paid mostly attention to differential outcome metrics, i.e., those which measure fairness as gaps in classification rates across groups~\cite{deFreitasPereira2020FairnessIB,Gong2019JointlyDF,grother2022face,Howard2022EvaluatingPF}. Compared to these, they proposed a differential performance metric based on the distances between score distributions $z$. This approach presents the main advantage of being agnostic to the operational point selected, thus measuring the fairness of the overall system.

Concretely, the metric proposed in~\cite{kotwal2023fairness} leverages the Kullback-Leibler (KL) divergence as the basic distance measure among the distributions of each demographic group. The metric, known as the Distribution Fairness Index (DFI), spans values between $0$ and $1$, where a value close to the latter represents a fairer model. Formally, DFI is defined as follows (using the same notation as in~\cite{kotwal2023fairness}):


\begin{equation}\label{eq:z_mean}
    z_{D_\mathrm{mean}}=\frac{1}{K} \sum_{i=1}^K z_{D_i}
\end{equation}


\begin{equation}
    \textrm{DFI}_\mathrm{N}=1-\frac{1}{K\log_2K}\sum_{i=1}^K S_i
    \label{eq:dfi_standard}
\end{equation}


\noindent where $z_{D_{i}}$ is the combined (genuine + impostor) distribution for the demographic group $d_i$, normalized so that the curve area sums one, and $S_i$ is the KL divergence among $z_{D_{i}}$ and the mean distribution $z_{D_\mathrm{{mean}}}$. The formulation in Eq.~\ref{eq:dfi_standard} corresponds to a baseline definition of DFI (Normal, therefore $\mathrm{N}$), which is based on the average of the dissimilarities of all the demographic distributions from the mean distribution. Additionally, it is a common approach to measure the fairness of a system considering the group that is disfavoured the most, as it represents the worst-performing case. Thus, another formulation of DFI can be made, which only considers the distribution of the demographic group that diverges the most from the mean:


\begin{equation}
    \textrm{DFI}_\mathrm{E}=1-\frac{1}{\log_2K}\max(S_i)
    \label{eq:dfi_extreme}
\end{equation}

Apart from DFI, two interesting differential outcome metrics are highlighted in the latest NIST FRVT report on demographic differentials measurement~\cite{grother2022face}. The first of them is an updated version of the Inequity metric, which computes the ratio between maximum and minimum FMR/FNMR values across demographic groups. As noted in~\cite{grother2022face}, using the minimum values is not robust in general setups, thus a better measurement can be obtained by including in the ratio the geometric mean of FMR/FNMR across groups, instead of the minimum. Thus, the modified Inequity metric is formulated as follows:

\begin{equation}
         \textrm{IN}_\mathrm{FMR} = \frac{\max_{d_i}\mathrm{FMR}(\tau)}{\mathrm{FMR}_\mathrm{geom}}\label{eq:inequity_fmr}
\end{equation}  

 \begin{equation}
         \textrm{IN}_\mathrm{FNMR} = \frac{\max_{d_i}\mathrm{FNMR}(\tau)}{\mathrm{FNMR}_\mathrm{geom}}\label{eq:inequity_fnmr}
    \end{equation}   

In addition to the previous metric, the NIST's report proposes as well the use of the GARBE metric to measure fairness in terms of both FMR and FNMR~\cite{howard2019effect}. This metric is inspired by the Gini coefficient, a commonly used measurement of income disparity, and is formulated as follows:

\begin{equation}
    \textrm{GARBE}_\mathrm{FMR} = \frac{\sum_{i}\sum_{j}|\mathrm{FMR}_{d_i}(\tau) - \mathrm{FMR}_{d_j}(\tau)|}{2K^2\,\mathrm{FMR}_\mathrm{arith}}
\end{equation}

\begin{equation}
    \textrm{GARBE}_\mathrm{FNMR} = \frac{\sum_{i}\sum_{j}|\mathrm{FNMR}_{d_i}(\tau) - \mathrm{FNMR}_{d_j}(\tau)|}{2K^2\,\mathrm{FNMR}_\mathrm{arith}}
\end{equation}

\noindent where $\mathrm{FMR}_\mathrm{arith}$ and $\mathrm{FNMR}_\mathrm{arith}$ representarithmetic means of each demographic group considered. The Gini metric yields values on the interval $[0, 1]$, with high values being associated with unfair systems. Higher values are a sign of unfairness as well for the Inequity metric. For both Inequity and GARBE metrics, an operational point needs to be selected. In its evaluations, NIST fixes the operational point as that for which the systems give an overall FMR of $0.0003$. Then, FMR and FNMR values for each demographic group can be computed and aggregated using any of the previous metrics. While both of them can be further aggregated into a single value, having a separate value for FMR and FNMR allows us to analyze the fairness with regard to different kinds of errors, i.e., whether the model exhibits more bias in the genuine distribution (FNMR) or in the impostor distribution (FMR).

\subsection{Proposed Metric: Comprehensive Equity Index (CEI)}
\label{sec:CEI}

In this section, we present an extension of the metric of~\cite{kotwal2023fairness} to measure fairness. Our proposal tries to keep the benefits of performance-based metrics while integrating the error-based perspective of differential outcome metrics. With this balance, we are not only aiming to measure the model's bias but also to consider how competitive the recognition system is, a relevant aspect in systems with very small error rates.

By examining the evaluation of high-performance models (e.g., those presented to NIST FRTE) with the DFI metric on state-of-the-art datasets, such as RFW~\cite{Wang2018RacialFI} or BUPT-B~\cite{wang2020mitigating}, we noticed that error rates associated to demographic biases are not captured with the cited metric. We hypothesize that, since DFI uses the entire distribution to measure fairness \textit{i)} the tail has a little relevance in the computation and \textit{ii)} genuine and impostor distributions are treated as a whole, hence hindering the assessment of any bias present in either of them. In comparison, differential outcome metrics such as GARBE~\cite{howard2019effect} or Inequity~\cite{grother2022face} can capture these biases, since the selection of an operational point directly focuses the evaluation on the tails of the distributions. However, measuring fairness for a concrete operational point presents some drawbacks. First, the demographic bias underlying the core of the biometric system is not captured at all, so information about the rest of the distribution is lost. Second, by considering only the tail of the distribution, the performance is measured in a lower percentage of samples than when using the entire curve. Thus, outcome differences could be due to reasons beyond demographic attributes, for instance, image resolution, brightness, or pose covariates.

We aim to overcome the aforementioned shortcoming by presenting a new fairness measure built on the proposal of Kotwal and Marcel~\cite{kotwal2023fairness}. Specifically, our objective is to have a metric that is both threshold-agnostic and able to measure bias in genuine and impostor distributions independently while properly accounting for the tails, i.e., where errors occur. We introduce here the Comprehensive Equity Index (CEI). For every demographic group, the CEI first splits each distribution (i.e., genuine or impostor) into two groups based on a given percentile $P_s$ (i.e., score threshold $s$ corresponding to certain accumulated probability $P$), dividing the tail from the rest of the distribution (referred to as center distribution from now on). The intuition here is to have independent components so we can assign them different weights when computing fairness. Once the distribution is split, we can compute a score $S_i^{\prime}$ (dissimilarity score as we are using distance measures) between a demographic distribution and the mean distribution as follows:


\begin{equation}
    S^{\prime}_{i}(P_s) = w_{t} \cdot D_\mathrm{KL}(z_{D_i}^t || z_{D_\mathrm{mean}}^t) +  w_{c} \cdot D_\mathrm{KL}(z_{D_i}^c || z_{D_\mathrm{mean}}^c)
    \label{eq:s_prime}
\end{equation}

\noindent where $z_{D_i}^t$ and $z_{D_i}^c$ are respectively the tail and center distributions from $z_{D_i}$, $z_{D_\mathrm{mean}}^t$ and $z_{D_\mathrm{mean}}^c$ refer to mean distributions as defined in Eq.~\ref{eq:z_mean}, and $w_{t}$ and $w_{c}$ are manually-tuned weights controlling the trade-off between the relevance of each part in the similarity score. The term in Eq.~\ref{eq:s_prime} is computed for each demographic group, then the CEI is calculated in a similar way as the DFI, having Normal and Extreme variants:

\begin{equation}
    \textrm{CEI}_\mathrm{N} (P_s) = 1-\frac{1}{K\log_2K}\sum_{i=1}^K S^{\prime}_i
    \label{eq:cei_standard}
\end{equation}

\begin{equation}
    \textrm{CEI}_\mathrm{E} (P_s) = 1-\frac{1}{\log_2K}\max(S^{\prime}_i)
    \label{eq:cei_extreme}
\end{equation}

Both proposed metrics $\textrm{CEI}_\mathrm{N}$ and $\textrm{CEI}_\mathrm{E}$ are on the interval $[0, 1]$, with a higher value being associated with a fairer model.

\section{Material and Methods}
\label{sec:materials}
\subsection{Models and Databases}
\label{sec:databases}

For the present work, we have trained two face recognition models from scratch for face recognition. The models were trained with a margin-based loss, i.e., CosFace~\cite{Wang2018CosFaceLM}, on the WebFace database~\cite{Zhu2021WebFace260MAB}, which contains $260$M images from $4$M identities. The database includes images from $7$ different race groups, with more than half of the identities being Caucasian. Similarly to the models evaluated in~\cite{Zhu2021WebFace260MAB}, we assessed the performance of the trained models on IJB-C~\cite{Maze2018158IJBC}. These models will be used later in our experiments:

\begin{itemize}
    \item \textbf{ResNet-100}~\cite{He2015DeepRL}. The ResNet architecture is one of the most famous convolutional models of the last decade. Here, we have used the architecture with 101 convolutional layers. The trained ResNet-100 model exhibits a FNMR@FMR=1e-5 of 0.0407.
    \item \textbf{Propietary Model}. A commercial model submitted to the NIST FRTE $1$:$1$ with a FNMR@FMR=0.0003 of 0.0058. When evaluating this model on the IJB-C~\cite{Maze2018158IJBC} dataset, we obtained FNMR@FMR=1e-5 of 0.037.
\end{itemize}

Throughout the experiments carried out in the present work, we will use the following publicly available databases: MORPH~\cite{Ricanek2006MORPH,2013PTomeFSI_FacialRegions}, RFW~\cite{Wang2018RacialFI}, and BUPT-B~\cite{wang2020mitigating}. All three databases include demographic labels with the gender and ethnicity of each subject. In addition to the aforementioned databases, we have used in this work a synthetic database recently released for the FRCSyn Challenge~\cite{melzi2024frcsyn} with realistic conditions and controllable demographics.

\section{Experiments}
\label{sec:experiments}

In this section, we present different experimental scenarios in which we show the usefulness of the proposed metric to measure the (un)fairness of high-performing face recognition models. In Section~\ref{sec:experiments_synthetic_dis} we present a toy scenario to elaborate on and numerically assess the advantages of the presented metric in comparison to existing metrics. Finally, experiments on real images are conducted in Section~\ref{sec:experiments_real}, where we evaluate a high-performance industry model and compare our proposed metric with existing methods.

\subsection{Synthetically-Generated Distributions}
\label{sec:experiments_synthetic_dis}



\begin{figure}[t!]
    \centering
    \includegraphics[width=\textwidth]{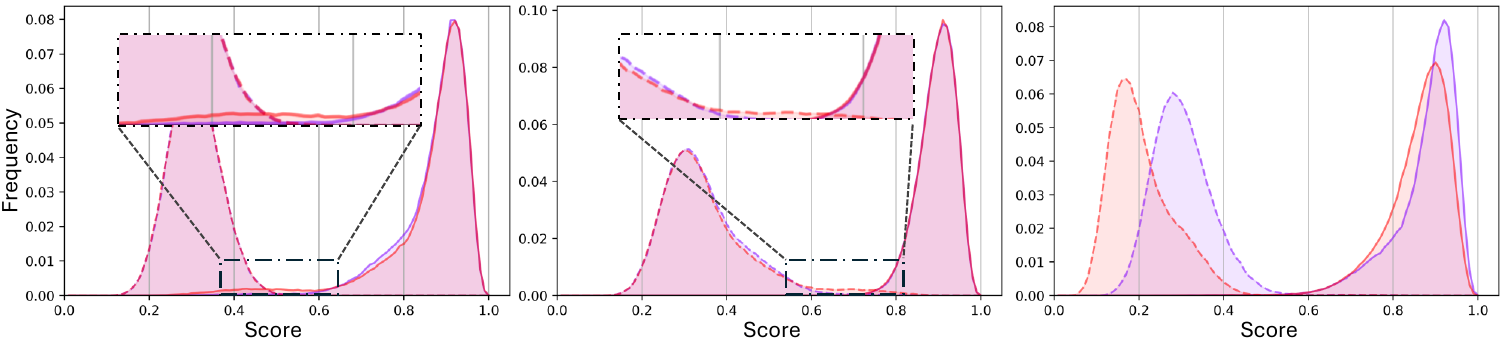}
    \caption{Genuine and impostor synthetically-generated similarity score distributions, in different scenarios: (Left) Biased Genuine distribution tail (BG); (Center) Biased Impostor distribution tail (BI); and (Right) Biased genuine-impostor distribution Center (BC).}
    \label{fig:syntheticcases2}
\end{figure}

In the following, we present experiments on synthetically-generated similarity score distributions, simulating the performance of a competitive model. Three scenarios are considered (see Fig.~\ref{fig:syntheticcases2}). First, we manipulated the left tail of the genuine distribution, i.e., the right distribution in Fig.~\ref{fig:syntheticcases2} (left), to increase the false rejections in that region. We have called this scenario Biased Genuine distribution tail (BG). This name is given because in the overlapping region between the two distributions (genuine and impostor), the genuine tail is forced (biased) to have an atypically high probability (considering as typical a rapid decrease similar in nature to a normal distribution tail, e.g., as shown in the impostor distribution).\footnote{In a general sense, bias in machine learning can be considered a systematic error that occurs in a model due to incorrect assumptions in the machine learning process. Technically, we can define bias as the error between the model behavior and the ground truth. In practical terms, measuring that error will normally mean measuring differences between score distributions, as done in the present paper for the particular case of systematic demographic differences in biometric systems~\cite{alcala2023measuring}.} The second scenario is similarly created for a Biased Impostor distribution tail (BI). Finally, in the third scenario, both distributions (genuine and impostor) have similar probabilities in their tails, but their centers are shifted. We have called this scenario Biased genuine-impostor distribution Centers (BC). The first two scenarios are expected to be well captured by the IN\textsubscript{FMR} (IN\textsubscript{FNMR}) and GARBE\textsubscript{FMR} (GARBE\textsubscript{FNMR}) metrics, as changes in the tail are more relevant here, whereas the distribution changes introduced in the third scenario will, in principle, be better captured using both DFI\textsubscript{N} and DFI\textsubscript{E} metrics, as the distribution tails in that case are similar and hence present an identical error rate.

\begin{table}[t!]
    \caption{Values of the DFI and NIST-related metrics and the proposed CEI\textsubscript{N} and CEI\textsubscript{E} on the simulated scenario.}\label{table:sota_metrics_synthetic}
    \centering
    \setlength{\tabcolsep}{6pt}
    \begin{tabular}{l|ccc|}
        \hline
         & BG & BI & BC \\
        \hline
        DFI\textsubscript{N}~\cite{kotwal2023fairness} & $0.9983$ &$0.9974$ & $0.8361$ \\
        DFI\textsubscript{E}~\cite{kotwal2023fairness} & $0.9982$ &$0.9970$ & $0.8112$ \\
        \hline
        GARBE\textsubscript{FMR}~\cite{howard2019effect} & $0.0050$ & $0.2950$ & $0.0433$ \\
        GARBE\textsubscript{FNMR}~\cite{howard2019effect} & $0.3326$ & $0.0025$ & $0.0208$ \\
        \hline
        IN\textsubscript{FMR}~\cite{grother2022face} & $1.1249$ & $2.0989$ & $1.0697$ \\
        IN\textsubscript{FNMR}~\cite{grother2022face} & $2.2331$ & $1.0037$ & $1.0416$ \\
        \hline
        CEI\textsubscript{N\textsubscript{Genuine}} [\textbf{ours}] & $0.5678$ & $0.9991$ & $0.9919$ \\
        CEI\textsubscript{N\textsubscript{Impostor}} [\textbf{ours}] & $0.9992$ & $0.6223$ & $0.3767$ \\
        CEI\textsubscript{E\textsubscript{Genuine}} [\textbf{ours}] & $0.4714$& $0.9990$ & $0.9916$ \\
        CEI\textsubscript{E\textsubscript{Imporstor}} [\textbf{ours}] & $0.9992$ & $0.5372$ & $0.2986$ \\
        \hline
        \end{tabular}
\end{table}

The evaluation presented in Table \ref{table:sota_metrics_synthetic} confirms the initial intuition. On one hand, both variants of the DFI metric are not able to detect any bias in those cases where the differences are found in the distribution tails (BG and BI), but the GARBE and IN metrics seem to capture those differences. We hypothesize that DFI\textsubscript{N} and DFI\textsubscript{E} are not being able to capture the generated bias because of: \textit{i}) the minor impact that differences in the distribution tail have compared to the center of the distribution, and \textit{ii}) the metric using the genuine + impostor distribution as a whole, ignoring particular differences found in each one. On the other hand, in the third scenario, it can be observed that the GARBE and IN metrics are not able to capture any differences, whereas both variants of the DFI seem to be more sensitive to distribution displacements. As the NIST-related metrics need the differences to be related to the performance instead of to the shape of the curve, this does not manifest in this case.






\begin{table}[t!]
    \caption{Values of CEI\textsubscript{N} on three synthetically generated cases: \textit{i)} BG ,\textit{ii)} BI, and \textit{iii)} BC. We evaluate each case using three different percentiles ($75$, $90$, $95$) and three different weight sets (i.e., \textbf{w}\textsubscript{1}=(0.2,0.8), \textbf{w}\textsubscript{2}=(0.5,0.5), and \textbf{w}\textsubscript{3}=(0.8,0.2)) for the tail and center of the distributions, respectively.}\label{table:cei_N_synthetic}
    \centering
    \setlength{\tabcolsep}{6pt}
    \begin{tabular}{cl|ccc|ccc|}
        \cline{3-8}
         && \multicolumn{3}{c|}{Genuine} & \multicolumn{3}{c|}{Impostor} \\
         && BG & BI & BC & BG & BI & BC \\
        \hline
        \multirow{3}{*}{\begin{sideways}$P\textsubscript{75}$\end{sideways}}&\textbf{w}\textsubscript{1} & $0.9897$ & $1.0000$ & $0.9110$ & $0.9999$ & $0.9908$ & $0.4008$ \\
        &\textbf{w}\textsubscript{2} & $0.9757$ & $0.9999$ & $0.8990$ & $0.9999$ & $0.9803$ & $0.4083$ \\
        &\textbf{w}\textsubscript{3} & $0.9617$ & $0.9998$ & $0.8869$ & $0.9998$ & $0.9698$ & $0.4158$ \\
        \hline
        \multirow{3}{*}{\begin{sideways}$P\textsubscript{90}$\end{sideways}}&\textbf{w}\textsubscript{1} & $0.9438$ & $0.9997$ & $0.9726$ & $0.9995$ & $0.9481$ & $0.5319$ \\
        &\textbf{w}\textsubscript{2} & $0.8787$ & $0.9994$ & $0.9757$ & $0.9992$ & $0.8901$ & $0.4580$ \\
        &\textbf{w}\textsubscript{3} & $0.8136$ & $0.9991$ & $0.9787$ & $0.9989$ & $0.8320$ & $0.3841$ \\
        \hline
        \multirow{3}{*}{\begin{sideways}$P\textsubscript{95}$\end{sideways}}&\textbf{w}\textsubscript{1} & $0.8791$ & $0.9998$ & $0.9804$ & $0.9997$ & $0.8936$ & $0.5703$ \\
        &\textbf{w}\textsubscript{2} & $0.7235$ & $0.9995$ & $0.9862$ & $0.9995$ & $0.7580$ & $0.4735$ \\
        &\textbf{w}\textsubscript{3} & $0.5678$ & $0.9991$ & $0.9919$ & $0.9992$ & $0.6223$ & $0.3767$ \\
        \hline
    \end{tabular}
\end{table}

To assess our proposed metric, we evaluated those three scenarios using different configurations of the proposed CEI metric with the normal variant, CEI\textsubscript{N}. We have conducted experiments combining percentile values of $0.75$, $0.90$, and $0.95$ and weight values of $(w\textsubscript{tail}, w\textsubscript{center})=\{(0.2, 0.8)\}, (0.5, 0.5), (0.8, 0.2)\}$. The results are shown in Table~\ref{table:cei_N_synthetic}. For the first two scenarios (BG and BI) of Fig.~\ref{fig:syntheticcases2}, it is observed that when configuring our metric to give more importance to the distribution tail (both using high percentile values and high $w\textsubscript{tail}$ proportions), our metric is able to detect the introduced bias (i.e., the CEI\textsubscript{N} value decreases) in each the genuine (for the BG scenario) and impostor distribution (for the BI scenario). (Note that the metric diverting from 1 means that the bias introduced between the two evaluated scenarios with/without bias is properly noticed.) For the last scenario, BC, we observe the biggest decrease in CEI\textsubscript{N} (i.e., largest bias detected) for the impostor distribution, as expected given the BC setup considered (see Fig. 1 right, where we can see that the bias introduced makes more different the impostor distributions in comparison to the genuine ones). Therefore, the proposed metric is able to detect the bias in all three presented cases, regardless of the weight parameters used. This is a desired behavior not observed with any of the other metrics in the literature. Thus we conclude that our proposed CEI has the potential to overcome some of the weaknesses of the original DFI. However, this needs to be assessed in real-world scenarios.

\subsection{Evaluation in Real Scenarios}
\label{sec:experiments_real}


\begin{figure}[t!]
    \centering
    \includegraphics[width=\textwidth]{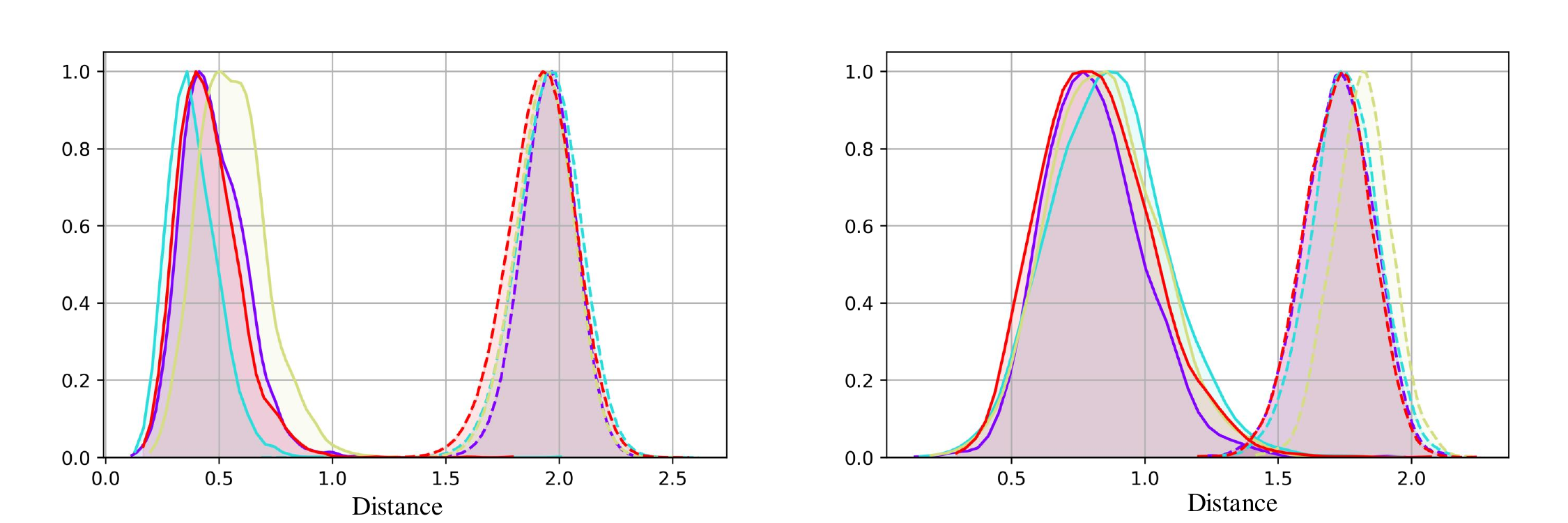}
    \caption{Genuine (continuous line) and impostor (dashed line) distributions for ResNet-100~\cite{He2015DeepRL} model in MORPH~\cite{Ricanek2006MORPH,2013PTomeFSI_FacialRegions} (Left) and RFW~\cite{Wang2018RacialFI} (Right) datasets. The x-axis shows the Euclidean distance between two images. Thus the genuine distributions are on the left and the impostor on the right. Each demographic group is represented by a different color.}
    \label{fig:real_cases_r100}
\end{figure}

\begin{table}[t!]
    \caption{Values of CEI\textsubscript{N} metric using the ResNet-100~\cite{He2015DeepRL} model on MORPH~\cite{Ricanek2006MORPH,2013PTomeFSI_FacialRegions}, RFW~\cite{Wang2018RacialFI}, BUPT-B~\cite{wang2020mitigating}, and the FRCSyn database~\cite{melzi2024frcsyn}. We evaluate using three different percentiles ($75$, $90$, $95$) and two weight sets (i.e., \textbf{w}\textsubscript{2}= (0.5,0.5), and \textbf{w}\textsubscript{3}= (0.8,0.2)) for the tail and center of the distributions, respectively.}\label{table:cei_N_real}
    \centering
    \setlength{\tabcolsep}{3pt}
    \begin{tabular}{cl|cccc|cccc|}
        \cline{3-10}
         && \multicolumn{4}{c|}{Genuine} & \multicolumn{4}{c|}{Impostor} \\
         && MORPH & RFW & BUPT-B & FRCSyn & MORPH & RFW & BUPT-B & FRCSyn\\
        \hline
        \multirow{2}{*}{\begin{sideways}p\textsubscript{75}\end{sideways}}&\textbf{w\textsubscript{2}} & $0.9759$ & $0.9318$ & $0.9192$ & $0.9515$ & $0.9492$ & $0.8151$ & $0.9670$ & $0.9316$\\
        &\textbf{w\textsubscript{3}} & $0.9614$ & $0.9073$ & $0.8951$ & $0.9343$ & $0.9368$ & $0.7556$ & $0.9532$ & $0.9138$\\
        \hline\hline
        \multirow{2}{*}{\begin{sideways}p\textsubscript{90}\end{sideways}}&\textbf{w\textsubscript{2}}& $0.9202$ & $0.8911$ & $0.8944$ & $0.9317$ & $0.9129$ & $0.7779$ & $0.9272$ & $0.9157$\\
        &\textbf{w\textsubscript{3}} & $0.8723$ & $0.8365$ & $0.847$ & $0.8988$ & $0.8741$ & $0.6804$ & $0.8907$ & $0.8798$\\
        \hline\hline
        \multirow{2}{*}{\begin{sideways}p\textsubscript{95}\end{sideways}}&\textbf{w\textsubscript{2}} & $0.8696$ & $0.8461$ & $0.8753$ & $0.9206$ & $0.8871$ & $0.7581$ & $0.9097$ & $0.9046$\\
        &\textbf{w\textsubscript{3}} & $0.7959$ & $0.7622$ & $0.8129$ & $0.8787$ & $0.8298$ & $0.6435$ & $0.8611$ & $0.8578$\\
        \hline
    \end{tabular}
\end{table}

\begin{table}[t!]
    \caption{Values of the DFI and NIST-related metrics, and the proposed CEI\textsubscript{N} and CEI\textsubscript{E} on the simulated scenario obtained on MORPH~\cite{Ricanek2006MORPH,2013PTomeFSI_FacialRegions}, RFW~\cite{Wang2018RacialFI}, BUPT-B~\cite{wang2020mitigating}, and the FRCSyn database~\cite{melzi2024frcsyn}.  The two CEI variants use a percentile of $95\%$ and weights ($w\textsubscript{tail}, w\textsubscript{center})=(0.8, 0.2)$.}\label{table:sota_metrics_real}
    \centering
    \setlength{\tabcolsep}{6pt}
    \subcaption*{ResNet-100~\cite{He2015DeepRL}}
        \begin{tabular}{l|cccc|}
        \cline{2-5}
         & MORPH & RFW & BUPT-B & FRCSyn\\
        \hline
        DFI\textsubscript{N}~\cite{kotwal2023fairness} & $0.9932$ & $0.9785$ & $0.9965$ & $0.9927$\\
        DFI\textsubscript{E}~\cite{kotwal2023fairness} & $0.9885$ & $0.9529$ & $0.9927$ & $0.9768$\\
        \hline
        GARBE\textsubscript{FMR}~\cite{howard2019effect} & $0.3762$ & $0.2885$ & $0.3289$ & $0.4631$\\
        GARBE\textsubscript{FNMR}~\cite{howard2019effect} & $0.1500$ & $0.1377$ & $0.2719$ & $0.0654$\\
        \hline
        IN\textsubscript{FMR}~\cite{grother2022face} & $2.9418$ & $1.8803$ & $2.2661$ & $4.3461$\\
        IN\textsubscript{FNMR}~\cite{grother2022face} & $1.6818$ & $1.3723$ & $2.0182$ & $1.1754$\\
        \hline
        CEI\textsubscript{N\textsubscript{Genuine}} [\textbf{ours}] & $0.7959$ & $0.7622$ & $0.8129$ & $0.8787$ \\
        CEI\textsubscript{N\textsubscript{Impostor}} [\textbf{ours}] & $0.8298$ & $0.6435$ & $0.8611$ & $0.8578$ \\
        CEI\textsubscript{E\textsubscript{Genuine}} [\textbf{ours}] & $0.5425$ & $0.6724$ & $0.6725$ & $0.6717$ \\
        CEI\textsubscript{E\textsubscript{Imporstor}} [\textbf{ours}] & $0.6797$ & $0.394$ & $0.7973$ & $0.6989$\\
        \hline
        \end{tabular}
    \subcaption*{Propietary Model}
        \begin{tabular}{l|cccc|}
        \cline{2-5}
         & MORPH & RFW & BUPT-B & FRCSyn\\
        \hline
        DFI\textsubscript{N}~\cite{kotwal2023fairness} & $0.9933$ & $0.9818$ & $0.9983$ & $0.9906$\\
        DFI\textsubscript{E}~\cite{kotwal2023fairness} & $0.9873$ & $0.9647$ & $0.9818$ & $0.9662$\\
        \hline
        GARBE\textsubscript{FMR}~\cite{howard2019effect} & $0.2439$ & $0.2500$ & $0.3075$ & $0.4616$\\
        GARBE\textsubscript{FNMR}~\cite{howard2019effect} & $0.1500$ & $0.1941$ & $0.2873$ & $0.0693$\\
        \hline
        IN\textsubscript{FMR}~\cite{grother2022face} & $2.9410$ & $1.7965$ & $2.1286$ & $4.1876$\\
        IN\textsubscript{FNMR}~\cite{grother2022face} & $1.6818$ & $1.5635$ & $2.0803$ & $1.2038$\\
        \hline
        CEI\textsubscript{N\textsubscript{Genuine}} [\textbf{ours}] & $0.9056$ &$ 0.7624$ & $0.7831$ & $0.8686$ \\
        CEI\textsubscript{N\textsubscript{Impostor}} [\textbf{ours}] & $0.9135$ & $0.6744$ & $0.9001$ & $0.8467$\\
        CEI\textsubscript{E\textsubscript{Genuine}} [\textbf{ours}] & $0.7953$ & $0.6408$ & $0.6431$ & $0.6704$\\
        CEI\textsubscript{E\textsubscript{Imporstor}} [\textbf{ours}] & $0.8585$ & $0.4560$ & $0.8492$ & $0.6867$\\
        \hline
        \end{tabular}
\end{table}

In Fig.~\ref{fig:real_cases_r100} the genuine and impostor distributions for a ResNet-100~\cite{He2015DeepRL} model trained over the WebFace database~\cite{Zhu2021WebFace260MAB} for the MORPH~\cite{Ricanek2006MORPH,2013PTomeFSI_FacialRegions} and RFW~\cite{Wang2018RacialFI} datasets is depicted. For each of the datasets, each curve represents a demographic group based on the ethnicity. We have evaluated the normal variant of the CEI metric described in Eq.~\ref{eq:cei_standard} with different configurations. More concretely, we analyze its behavior using percentile values of $75\%$, $90\%$, $95\%$, and, based on the observations from Section~\ref{sec:experiments_synthetic_dis}, weight values of $(w\textsubscript{tail}, w\textsubscript{center})=\{(0.5, 0.5), (0.8, 0.2)\}$. Table~\ref{table:cei_N_real} it can be observed that, as the distribution tail receives more importance (i.e., using high percentile like $P\textsubscript{95}$ and a weight combination that prioritizes the tail like $\textbf{w}_3$). the metric value decreases, indicating that differences among demographic groups exist on those parts of the curves. In Table 3 in that configuration ($P\textsubscript{95}$ and $\textbf{w}\textsubscript{3}$) we also observe differences in the behavior of genuine and impostor distributions, e.g., MORPH~\cite{Ricanek2006MORPH,2013PTomeFSI_FacialRegions} and RFW~\cite{Wang2018RacialFI} have a larger difference in their CEI\textsubscript{N} score between the genuine and impostor distribution, meaning that the bias is different for each one. We have used the distance score distributions over those datasets and the ResNet-100~\cite{He2015DeepRL} model in figure ~\ref{fig:real_cases_r100} to confirm the existence of the differences captured by the CEI\textsubscript{N}.

The configuration using a percentile of $95\%$ and weights (textsubscript)=(0.8, 0.2) has been used for the two variants of the proposed CEI metric (CEI\textsubscript{N}) and the extreme variant described in Equation~\ref{eq:cei_extreme}, (CEI\textsubscript{E}) to compare them with other existing metrics (see Section ~\ref{sec:metrics}). The results are represented in Table~\ref{table:sota_metrics_real}. It is shown that the DFI-related metrics (DFI\textsubscript{N} and DFI\textsubscript{E}) are not able to capture any of the existing differences. As hypothesized before, this may be related to the fact that differences are mainly found in the distribution tails. Moreover, the DFI\textsubscript{N} and DFI\textsubscript{E} metrics do not separate the genuine and impostor distributions. It uses an aggregation of both distributions to compute the "fairness", provoking bias related to specific distributions not to be captured. That behavior is not observed with the NIST-related metrics, which is especially relevant in this scenario because the metric can detect potential differences between the demographic groups while providing more detailed information about the distribution (genuine or impostor) in which the difference is found. If we analyze the results obtained for both variants of the CEI metric, we find improvements w.r.t. the existing performance-based DFI metric, as it is able to better detect differences between demographic groups. Moreover, we observe that the proposed variant CEI\textsubscript{E} is more sensitive when measuring those demographic differences. Thus, the validity of both variants of the proposed metric is confirmed. It is shown to be able to capture existing differences while maintaining the strengths of the performance-based approach.

\section{Conclusions}
\label{sec:conclusions}
In this work, we follow up on previous efforts to measure "fairness" in biometric recognition systems by using a differential performance-based approach, dependent on the system score function. We have introduced a modification of a previous metric by adapting it into its application to real-world scenarios where the differences are found in the score distribution tails. The proposed metric, called Comprehensive Equity Index (CEI), has been shown to capture existing differences in the score distributions for different demographic groups when evaluating a high-performance Face Recognition (FR) system presented in the NIST FRVTE $1$:$1$ (with excellent results) in several state-of-the-art datasets.

Our proposal addresses previous weaknesses of differential performance metrics by parameterizing the relevance of the tail distribution differences for diverse demographic groups with a percentile selecting the tail and weights that give more or less importance to differences in that area of the distribution. Our proposed metric CEI also provides information on the bias encountered in each of the genuine and impostor distributions. This way, the metric can adapt to the distribution area where bias is desired to be studied. The proposed metric therefore overcomes observed deficiencies of previous metrics in real-world scenarios while preserving the benefits of the differential performance approach: it does not depend on concrete operational points and knowledge of the intrinsic behavior of the system, i.e., how the model represents biometric samples depending on its demographic attributes.

The introduced metric should be understood as a complement to other performance outcome-based metrics. Ours can detect differences in distributions, but this may not always be enough to determine whether a system is fair (or has bias), as that statement is dependent on the definition of fairness (or bias) and the concrete use case~\cite{2022_AI_SensitiveLoss_IS}. We propose to use the CEI as an index to detect differences in high-performance model distributions together with other performance metrics such as FMR, FNMR, and outcome differential-based indexes to have a wider view of the biases of the system in terms of the demographic group.

Future work includes continued investigation on data-efficient and cost-effective bias detection and evaluation methods looking both at models internals~\cite{2022_SafeAI_IFBiD_Serna,serna2021insidebias} and  outputs~\cite{alcala2023measuring}, symbolic methods to analyze biases~\cite{2023_ECAIw_LFIT-XAI_Tello}, and exploitation of LLMs to better assess biometric systems~\cite{ivan2024gpt} including bias evaluation.

\subsubsection{Acknowledgements}


This paper has been financed by the Government of Navarre within Industrial Doctorates 2022, the company Veridas\footnote{\url{https://veridas.com/en/}}, Cátedra ENIA UAM-VERIDAS en IA Responsible (NextGenerationEU PRTR TSI-100927-2023-2), and project BBforTAI (PID2021-127641OB-I00
MICINN/FEDER).

%
%
%
\bibliographystyle{splncs04}
\bibliography{refs}
\end{document}